\pdfoutput=1 

\documentclass[a4paper,fleqn]{cas-sc}

\usepackage[numbers]{natbib}
\usepackage{graphicx}
\usepackage{float}
\usepackage{url,hyperref,cleveref}
\usepackage{todonotes}
\usepackage{algorithm}
\usepackage{algpseudocode}
\usepackage{algorithmicx}
\usepackage{subcaption}
\usepackage{algorithm} 
\usepackage{listings}

\begin{document}
\let\WriteBookmarks\relax
\def\floatpagepagefraction{1}
\def\textpagefraction{.001}

\algdef{SE}[SUBALG]{Indent}{EndIndent}{}{\algorithmicend\ }%
\algtext*{Indent}
\algtext*{EndIndent}

\lstset{ %
	language=Java,                
	basicstyle=\footnotesize,       
	numbers=left,                   
	numberstyle=\footnotesize,      
	stepnumber=1,                   
	numbersep=5pt,                  
	backgroundcolor=\color{white},  
	showspaces=false,               
	showstringspaces=false,         
	showtabs=false,                 
	frame=single,           
	tabsize=2,          
	captionpos=b,           
	breaklines=true,        
	breakatwhitespace=false,    
	escapeinside={\%*}{*)}          
}

\shorttitle{MyDigiTwin: A Privacy-Preserving Framework for Personalized Cardiovascular Risk Prediction and Scenario Exploration}

\shortauthors{Cadavid et~al.}

\title [mode = title]{MyDigiTwin: A Privacy-Preserving Framework for Personalized Cardiovascular Risk Prediction and Scenario Exploration}                      



%
\author[1]{H\'{e}ctor Cadavid}[type=editor,
                        auid = 000, bioid = 1, orcid = 0000-0003-4965-4243]
\ead{h.cadavid@esciencecenter.nl}
\credit{Conceptualization of this study, Methodology, Software}

\affiliation[1]{organization={Netherlands eScience Center},
    addressline={Matrix THREE, Science Park 402}, 
    city={Amsterdam},
    postcode={1098 XH}, 
    country={The Netherlands}}

\affiliation[2]{
    organization={Department of Radiology \& Nuclear Medicine, Erasmus MC University Medical Center Rotterdam},
    addressline={Dr. Molewaterplein 40}, 
    city={Rotterdam},
    postcode={3015 GD}, 
    country={The Netherlands}}

\affiliation[3]{
    organization={Institute of Cardiovascular Science, Faculty of Population Health Sciences, University College London},
    addressline={62 Huntley St}, 
    city={London},
    postcode={WC1E 6DD}, 
    country={United Kingdom}}

\affiliation[4]{
    organization={Department of Cardiology, Amsterdam Cardiovascular Science, Amsterdam University Medical Centers, University of Amsterdam},
    addressline={Meibergdreef 9}, 
    city={Amsterdam},
    postcode={1105 AZ}, 
    country={The Netherlands}}

\affiliation[5]{
    organization={Department of Epidemiology, Erasmus MC University Medical Center Rotterdam},
    addressline={Dr. Molewaterplein 40}, 
    city={Rotterdam},
    postcode={3015 GD}, 
    country={The Netherlands}}

\affiliation[6]{
    organization={Department of Cardiology, University Medical Center Utrecht},
    addressline={Heidelberglaan 100}, 
    city={Utrecht},
    postcode={3584 CX}, 
    country={The Netherlands}}

\author[2]{Hyunho Mo}[type=editor,
                        auid = 000, bioid = 1, orcid = 0000-0002-6497-2250]
\cormark[1]
\fnmark[1]
\ead{h.mo@erasmusmc.nl}
\author[6]{Bauke Arends}[type=editor,
                        auid = 000, bioid = 1, orcid = 0009-0009-7494-4542]

\author[3,4]{Katarzyna Dziopa}[type=editor,
                        auid = 000, bioid = 1, orcid = 0000-0001-9497-0208]


\credit{Data curation, Writing - Original draft preparation}

\author[2]{Esther E. Bron}[type=editor,
                        auid = 000, bioid = 1, orcid = 0000-0002-5778-9263]

\author[2,5]{Daniel Bos}[type=editor,
                        auid = 000, bioid = 1, orcid = 0000-0001-8979-2603]

\author[1]{Sonja Georgievska}[type=editor,
                        auid = 000, bioid = 1, orcid = 0000-0002-8094-4532]
\fnmark[2]

\author[6]{Pim {van der Harst}}[type=editor,
                        auid = 000, bioid = 1, orcid = 0000-0002-2713-686X]


\fnmark[2]

\fntext[1]{Corresponding author.}
\fntext[2]{These authors contributed equally to this work.}


\begin{abstract}
Cardiovascular disease (CVD) remains a leading cause of death, and primary prevention through personalized interventions is crucial. This paper introduces MyDigiTwin, a framework that integrates \textit{health digital twins} with \textit{personal health environments} to empower patients in exploring personalized health scenarios while ensuring data privacy. MyDigiTwin uses federated learning to train predictive models across distributed datasets without transferring raw data, and a novel data harmonization framework addresses semantic and format inconsistencies in health data. A proof-of-concept demonstrates the feasibility of harmonizing and using cohort data to train privacy-preserving CVD prediction models. This framework offers a scalable solution for proactive, personalized cardiovascular care and sets the stage for future applications in real-world healthcare settings.





\end{abstract}


\begin{highlights}
\item MyDigiTwin integrates \textit{health digital twins} with \textit{personal health environments}.
\item The \textit{personal health train} concept enables privacy-preserving, collaborative predictive modeling.
\item A novel data harmonization framework resolves semantic inconsistencies in health data.
\item Proof-of-concept demonstrates federated learning using harmonized cohort data for CVD prediction models.
\item MyDigiTwin offers a scalable solution for proactive, personalized cardiovascular care.
\end{highlights}


\maketitle

\section{Introduction}

Cardiovascular disease (CVD) remain a leading cause of deaths worldwide, with many diagnoses occurring only after the onset of symptoms. This reactive, symptom-driven approach may hinder timely interventions, emphasizing secondary prevention — managing conditions after they arise — over proactive primary prevention. Primary prevention, which involves identifying at-risk individuals before symptoms manifest and empowering them to act on modifiable risk factors, such as lifestyle and diet, has the potential to significantly reduce cardiovascular events and alleviate healthcare burdens~\cite{bakhit_cardiovascular_2024}. Achieving this, however, requires innovative frameworks that integrate personalized data, predictive models, and actionable insights to address the complexities of cardiovascular health management.

One promising innovation for advancing primary prevention is the concept of \textit{health digital twin}. Health digital twins provide virtual representations of patient-specific conditions, enabling personalized risk assessments, outcome predictions, and scenario exploration. By integrating predictive modeling with patient-specific data, this concept offer the potential to shift CVD prevention from a reactive approach to a proactive, data-driven strategy. However, implementing this type of digital twin at scale for CVD prevention poses several challenges~\cite{venkatesh2024health, coorey2022health, sel2024building}, among which:

\begin{itemize}
    \renewcommand{\labelitemi}{-} 
    \item \textit{Data integration and harmonization:} Health data is often distributed across various systems and formats, which requires a data unification procedure when attempting to integrate it into a cohesive predictive framework.
    \item \textit{Privacy and security:} It is important to ensure that sensitive patient data is handled in a privacy-conscious manner during model training and prediction.
    \item \textit{Scalable patient engagement:} Digital twins need to find ways to bridge the gap between generalized insights and more individualized, actionable feedback for patients.
\end{itemize}

To contribute to addressing these challenges, this paper introduces MyDigiTwin, a framework for building scalable and privacy-preserving cardiovascular health digital twins. MyDigiTwin leverages two key components:

\begin{itemize}
    \renewcommand{\labelitemi}{-} 
    \item \textit{Personal Health Environments:} Digital platforms designed to consolidate fragmented health records into a patient-accessible interface, empowering individuals to engage with their health data and explore personalized health scenarios.
    \item \textit{Personal Health Train:}
A distributed infrastructure that employs federated learning to build personalized health prediction models securely by bringing algorithms to the data sources. This approach facilitates collaborative research across institutions while ensuring compliance with privacy regulations.
\end{itemize}

A core feature of the MyDigiTwin framework is the ability to simulate hypothetical scenarios, such as the effects of lifestyle changes, through predictive models integrated within the \textit{personal health environments}. Effective scenario exploration relies on accurate predictive models, which ideally would be trained using diverse datasets from multiple institutions. However, privacy regulations often prevent the sharing of data between hospitals, presenting a common challenge in the development of  \textit{health digital twins}. To overcome this, MyDigiTwin leverages federated learning~\cite{mcmahan2017communication, kaissis2021federated}, which allows models to be trained collaboratively across institutions without the need to transfer sensitive raw data.

Additionally, the integration of data from \textit{personal health environments} and hospital systems is hindered by inconsistencies in data formats. To resolve this, MyDigiTwin introduces a data harmonization framework based in the \textit{Fast Healthcare Interoperability Resources} (FHIR)~\cite{HL7FHIR} standard. Specifically, it uses the same FHIR profile that these healthcare environments use for data exchange, facilitating interoperability and seamless integration of predictive capabilities.

To validate the proposed framework, this paper presents a proof-of-concept implementation. This includes the harmonization of the Lifelines~\cite{scholtens2015cohort} dataset, utilizing commonly used CVD predictors, and deploying it within a federated learning infrastructure to train a CVD prediction model with 10-year follow-up. Although this model is not ready for clinical application, it effectively demonstrates the feasibility of the MyDigiTwin framework in harmonizing data and enabling privacy-preserving predictive modeling.

The remainder of this paper is structured as follows. \Cref{sec:dutch-helth-care} provides context for the need of the MyDigiTwin framework by giving an overview of the \textit{personal health environments} in the Dutch healthcare system. \Cref{sec:architecture} details the architecture of the MyDigiTwin framework. \Cref{sec:harmonization} discusses the design and implementation of the data harmonization pipeline. \Cref{sec:poc} presents the proof-of-concept implementation as well as validation results, and \Cref{sec:idiscussion}  discusses the implications, challenges, and future directions of the framework. Finally, \Cref{sec:conclusions} concludes the paper with key findings and potential avenues for future research.

\section{Personal Health Environments  and the Dutch healthcare system}\label{sec:dutch-helth-care}

In recent years, the Dutch Ministry of Health, Welfare and Sport has launched several initiatives to improve the exchange of health information between patients and healthcare providers. These initiatives were complemented by the Dutch law on electronic electronic data exchange in healthcare (\textit{Wet electronische gegevensuitwisseling in de zorg}), which aims to facilitate digital information sharing across the healthcare system. Among the most notable developments is the \textit{personal health environment}, a digital platform that enables users to access their health data in a standardized format, independent of the healthcare provider~\cite{netherlands}. This platform addresses a long-standing challenge in the Dutch healthcare system, the fragmentation of patient health records. Historically, health data in the Netherlands has been scattered across various healthcare providers, such as hospitals, general practitioners, and pharmacies, resulting in fragmented records that hinder comprehensive health management. Patients often face difficulties in piecing together their complete medical data. These environments overcome this barrier by aggregating data from multiple sources into a single, user-friendly interface, empowering patients to take an active role in managing their health.

\begin{figure}[h]
	\centering
	\includegraphics[width=0.7\linewidth]{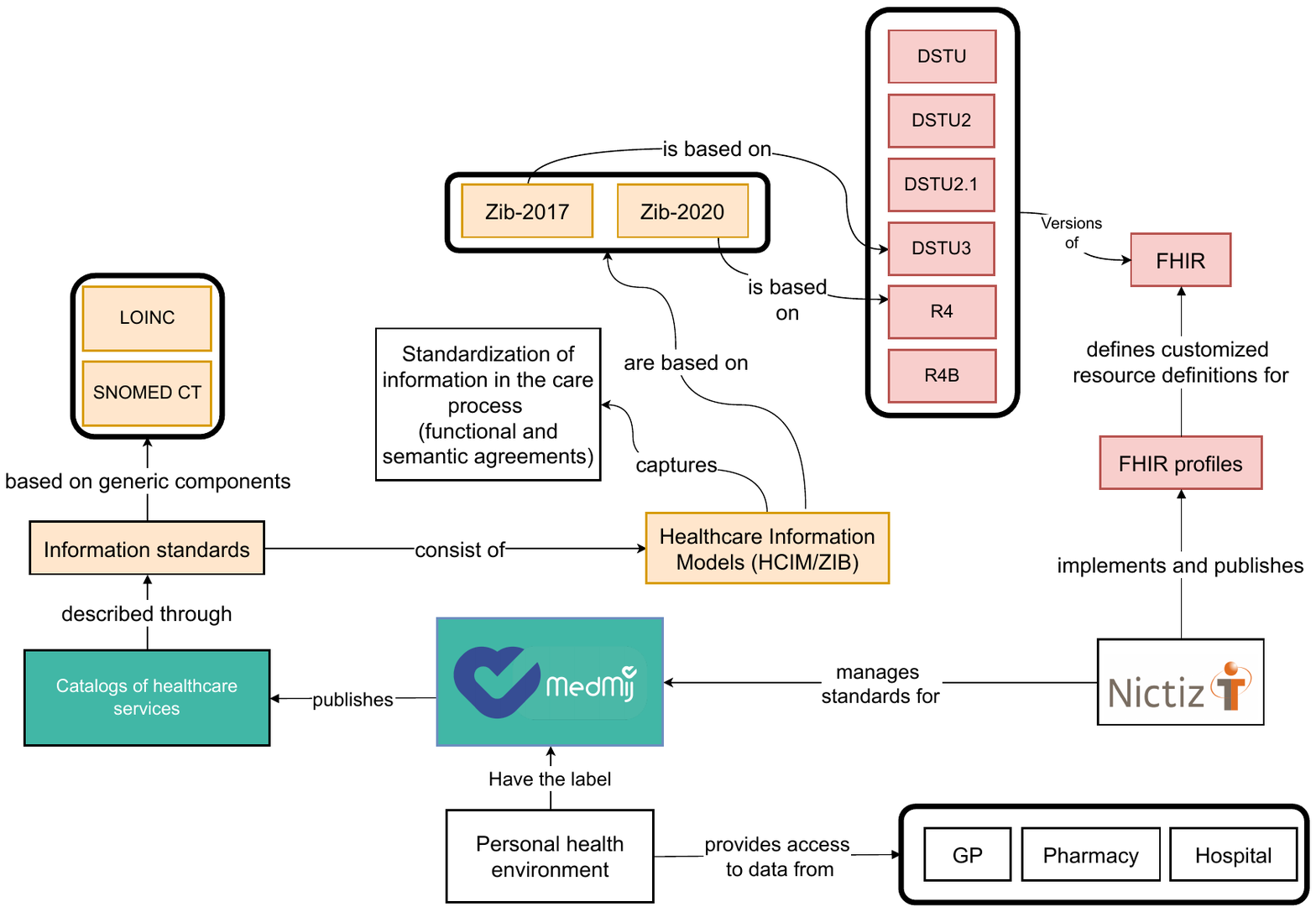}
	\caption{Overview of concepts related to personal health environments, the Dutch healthcare system, and the MyDigiTwin framework. Once awarded the MedMij label, a personal health environment is allowed to access data using the standardized information standards that are designed by MedMij and Nictiz.}
	\label{fig:PHE-concepts}
\end{figure}

The \textit{personal health environment} initiative aligns with the goals set out in the \textit{integral care agreement} (in Dutch: \textit{integraal zorgakkoord}), an agreement among key stakeholders in Dutch healthcare, which aims to provide all Dutch residents with access to a user-friendly, data-rich environment by 2025~\cite{ministerievws2022}. In addition to addressing the issue of fragmented health information, this concept also contributes to a broader commitment to promoting the free exchange of healthcare data, both nationally and at the European level, through frameworks such as the European Health Data Space~\cite{european_commission_european_2024}. 

To achieve these goals, the success of the \textit{personal health environments} rely on a robust framework for data standardization and interoperability. The MedMij foundation provides this essential framework, facilitating digital data exchange between care users and their healthcare providers (Figure~\ref{fig:PHE-concepts}). By adhering to MedMij’s criteria, healthcare providers can exchange data with any care user, ensuring the system remains both scalable and manageable~\cite{medmij_medmij_2024}. MedMij, in collaboration with \textit{personal health environment} suppliers, patient representatives, and other stakeholders, is responsible for designing the information standards – technical specifications that healthcare information systems must adhere to in order to facilitate seamless data exchange. These standards ensure that health data, such as blood pressure measurements and hemoglobin levels, are exchanged in a structured and consistent format; for example, ensuring that entities like blood pressure are always recorded and shared in the same measurement units (e.g., mmHg).

To promote interoperability, MedMij incorporates widely recognized components like Systematized Nomenclature of Medicine Clinical Terms (SNOMED CT) and Logical Observation Identifiers Names and Codes (LOINC) into its information standards~\cite{forrey_logical_1996, snomed_international_5-step_2024}. Additionally, it defines and publishes specific HL7 FHIR profiles to facilitate structured health data exchange. FHIR profiles are customizations of the base FHIR specification, tailored to meet the needs of particular contexts, such as specific healthcare domains. These profiles constrain the general-purpose FHIR resources by specifying which elements are required or optional, adding new elements, and defining terminology bindings to align with local requirements. In the MedMij framework, these adaptations ensure that the generic FHIR components are implemented consistently, leading to interoperability across \textit{health environments} and other healthcare systems. 

Although \textit{ personal health environments} represent a significant advance in health data access, their functionality is generally limited to displaying raw health records without offering interpretation or actionable insights. The following section elaborates on how MyDigiTwin builds upon these foundational capabilities.


\section{MyDigiTwin system architecture}\label{sec:architecture}

\begin{figure}[th]
	\centering
	\includegraphics[width=0.7\linewidth]{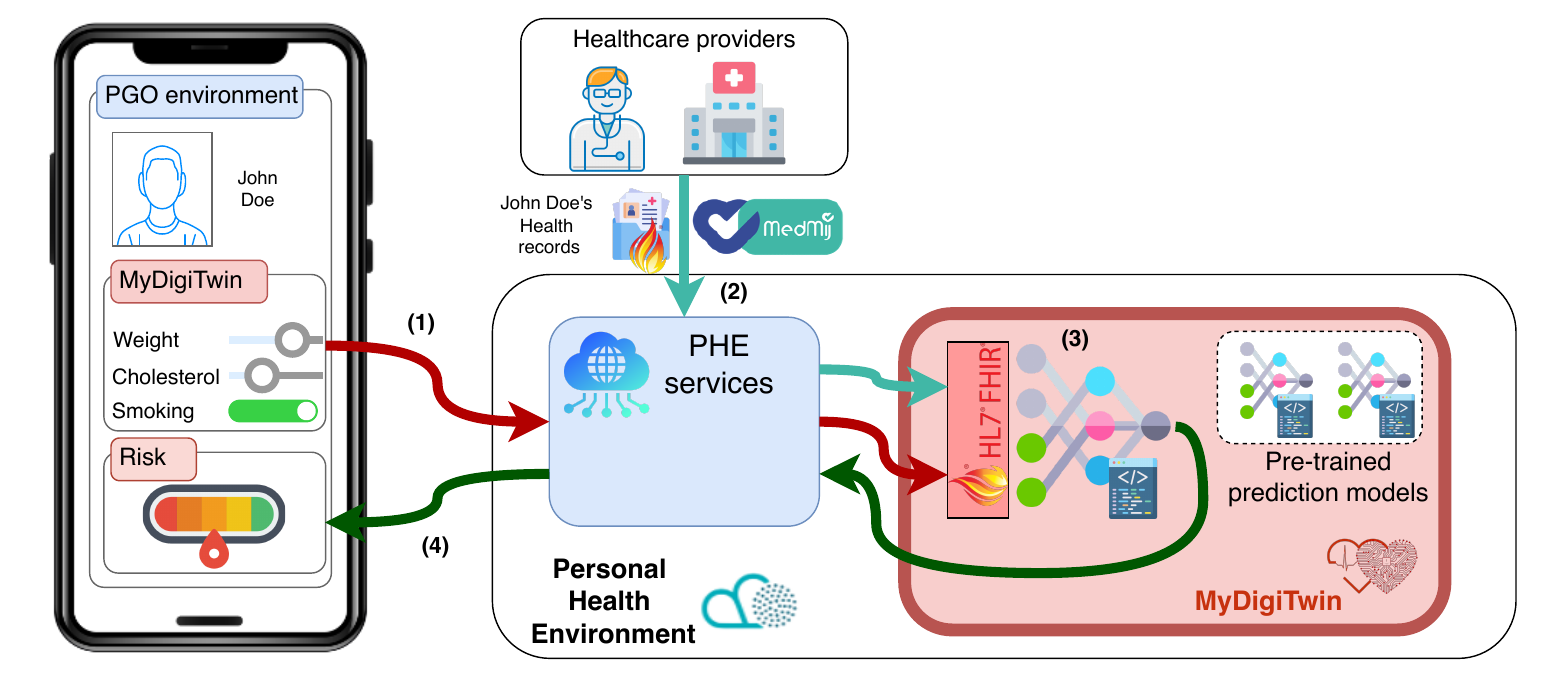}
	\caption{Overview of MyDigiTwin's end-user early detection and simulation services. (1) The end-user provides scenarios of health indicators for his/her DigiTwin to be simulated. (2) The \textit{personal health environment} (PGO) gets access to the end-user indicators available from (MedMij-compliant) health care providers. (3) The prediction model that fits better the end-user is selected from the set of pre-trained prediction  models. This model is fed with both modified and actual indicators based on the model's metadata -which defines its expected input-. (4) The risk estimation, calculated by the prediction model for the given scenario, is sent back to the end-user.}
	\label{fig:pgo-simulation-service}
\end{figure}

At its core, MyDigiTwin's architecture transforms raw health records into actionable insights by seamlessly integrating pre-trained cardiovascular disease (CVD) prediction models into the data flow of a \textit{personal health environment}. As illustrated in Fig.~\ref{fig:pgo-simulation-service}, the MyDigiTwin module --when running within one of these health environments-- feeds the indicators available in the patient's health records to these prediction models. It then sends back to the patient the models' output as personalized insights (e.g., risk assessments, recommendations, etc.). These models, which are developed, tested and refined by domain experts in a dedicated \textit{research environment}, can also be used to simulate lifestyle change scenarios by adjusting these input indicators, as depicted on the mobile user interface on the left part of Fig.~\ref{fig:pgo-simulation-service}. 

\begin{figure}[th]
	\centering
	\includegraphics[width=0.7\linewidth]{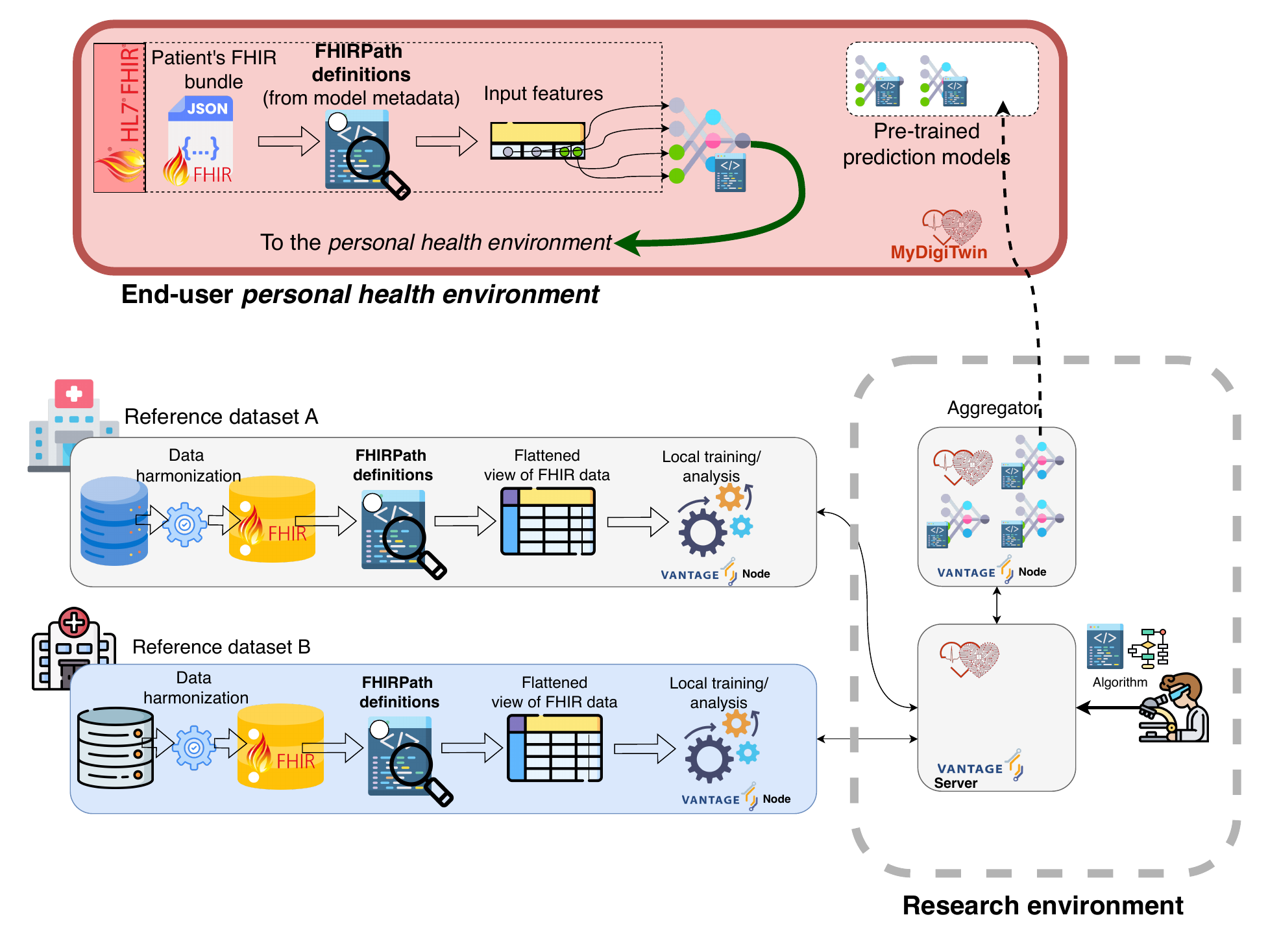}
	\caption{MyDigiTwin framework high-level system architecture. The organizations involved on \textit{research environment} (bottom left) agreed on which variables the vantage6 node will have access to through a set of FHIRPath expressions. A researcher request the execution of an algorithm for performing a federated analysis or model training (bottom right), also considering these ``canonical'' FHIRPath expressions. On the end-user environment (top), when the model is required for a given user prediction or simulation, the same FHIRPath expressions, that are part of the model metadata, are used to get (consistently with the training process) the indicators that will be used as input.}
	\label{fig:fhir-for-training-prediction-consistency}
\end{figure}

The aforementioned \textit{research environment}, one of the key components of the MyDigiTwin architecture (see Fig.~\ref{fig:fhir-for-training-prediction-consistency}), enables CVD researchers to perform experiments with the available big data reference datasets and to train the prediction models that will eventually be deployed on the \textit{personal health environment} depicted in Fig.~\ref{fig:pgo-simulation-service}. Considering the strict privacy regulations put in place by institutions to access their data for research purposes, this research infrastructure is based on the concept of \textit{Personal Health Train}~\cite{beyan2020distributed}. In a \textit{personal health train}, the sensitive data included in the reference data sets (the ``stations'') remain secure within their original location, and the algorithms (the ``trains'') travel to these data ``stations'' through a secure network (the ``tracks''). In this way, only aggregated data is sent back, for consolidation, when these algorithms are executed on each data ``station''. Vantage6~\cite{moncada2020vantage6}, a federated learning framework inspired in this concept, is used as the research infrastructure backbone. As illustrated in Fig.~\ref{fig:fhir-for-training-prediction-consistency}, this infrastructure comprises a server to orchestrate federated analysis / learning processes, a node to aggregate the results of the federated computations, and one additional node running on the premises where each reference dataset resides.

With this \textit{personal health train} setup, adding a new reference data set involves adding a new data ``station'', that is, setting up a new \textit{vantage6} node in the hospital or institute where the data set is located. However, since one of the key purposes of the federated infrastructure is to train prediction models on multiple reference datasets as if they were forming a single horizontally partitioned dataset~\cite{yang2020horizontal}, a significant challenge must first be addressed: ensuring consistency between datasets considering elements such as structure, terminology, and measurement units. Furthermore, it is also imperative to ensure consistency between these data sets and the health records used as input for the models when making predictions or simulating scenarios.

Given this, a data harmonization process, using the same FHIR profile used by the \textit{ personal health environments} described in Section~\ref{sec:dutch-helth-care} as the target format, was defined as the preliminary step to the integration of new data sets into the research infrastructure. Having all data stations with ZIB/FHIR-compliant data ensures that a federated model training can be performed using a clear and unambiguous specifications of the variables that will be used in the process. Once the model has been trained using the (federated) harmonized data and transferred to the \textit{personal health environment}, the same specifications can be used to consistently extract, from the patient's health records, the precise input (in terms of variables, measure units, etc.) required by the model for prediction purposes. 

The diagram in Fig.~\ref{fig:fhir-for-training-prediction-consistency} illustrates how the FHIRPath\footnote{FHIRPath is an expression language for navigating and extracting data from FHIR resources.} language is used in the system architecture to maintain consistency during the training process (in the research infrastructure), and when making patient-specific predictions (in the \textit{ personal health environment}). In the data stations of the organizations involved, a set of FHIRPath expressions is used to specify which specific properties of the harmonized FHIR dataset will be accessible by the federated learning infrastructure. Based on this specification, each data station uses the same tools~\cite{Cadavid_SQLonFHIRProjections} for generating a machine-learning-friendly flattened\footnote{Flattening is the process of converting hierarchical, nested data structures from FHIR resources into a two-dimensional, tabular format suitable for machine learning algorithms.} version of the data that a vantage6-compliant model training process can work with. The models resulting from this training process, which are eventually transferred to the \textit{personal health environments}, will include these FHIRPath expressions as part of their metadata\footnote{AI model metadata refers to information that describes key aspects of an AI model, including its name, version, architecture, training data, performance metrics, deployment environment, input/output definitions, etc.}. 

In this manner, when a patient requests a simulation, the MyDigiTwin module running on the health environment employs these FHIRPath expressions to precisely select and retrieve the necessary input values from the patient's health records.


\section{Harmonization of federated cohort studies}\label{sec:harmonization}

For MyDigiTwin's research environment, it is essential that the reference datasets are harmonized using the ZIB/FHIR profile. As previously described, this is not only because federated learning requires having data consistency across all participating nodes, but also because it ensures that the prediction models created in this environment are interoperable with the health records to which the \textit{ personal health environments} have access to. However, this data harmonization process presents unique complexities, as the research environment is not limited to a predefined collection of reference data sets. Instead, it is envisioned as a recurring process that must be replicated as new datasets are integrated into the research infrastructure over time. 

Harmonizing a single data set into the target FHIR format involves the development of a data pipeline with Extraction, Transformation, and Loading phases (ETL). \textit{Extraction} involves retrieving data from its original source, while \textit{Loading} focuses on inserting the transformed data into the target FHIR database. However, in this medical setting, the \textit{Transformation} phase stands out as the most intricate part of ETL due to its complexity. It involves an interdisciplinary collaboration process between pipeline developers, data experts who understand the underlying semantics of the source data, and medical domain experts who can properly define the mapping between the original data and the target format's semantics -- hereafter referred to as \textit{pairing rules}. Therefore, sScaling this process to multiple heterogeneous reference datasets in an efficient way, that is, making parts of the process reusable across datasets, becomes a significant challenge. This is particularly important considering that the unique characteristics of the overall framework limit the use of existing solutions to implement these harmonization pipelines. For instance, most existing data harmonization platforms are not designed to function in a federated setting, and the ones designed for it either do not support FHIR as the target format or are not designed to process cohort-study data~\cite{cadavid2024leveraging}. 

\begin{figure}[th]
	\centering
	\includegraphics[width=0.9\linewidth]{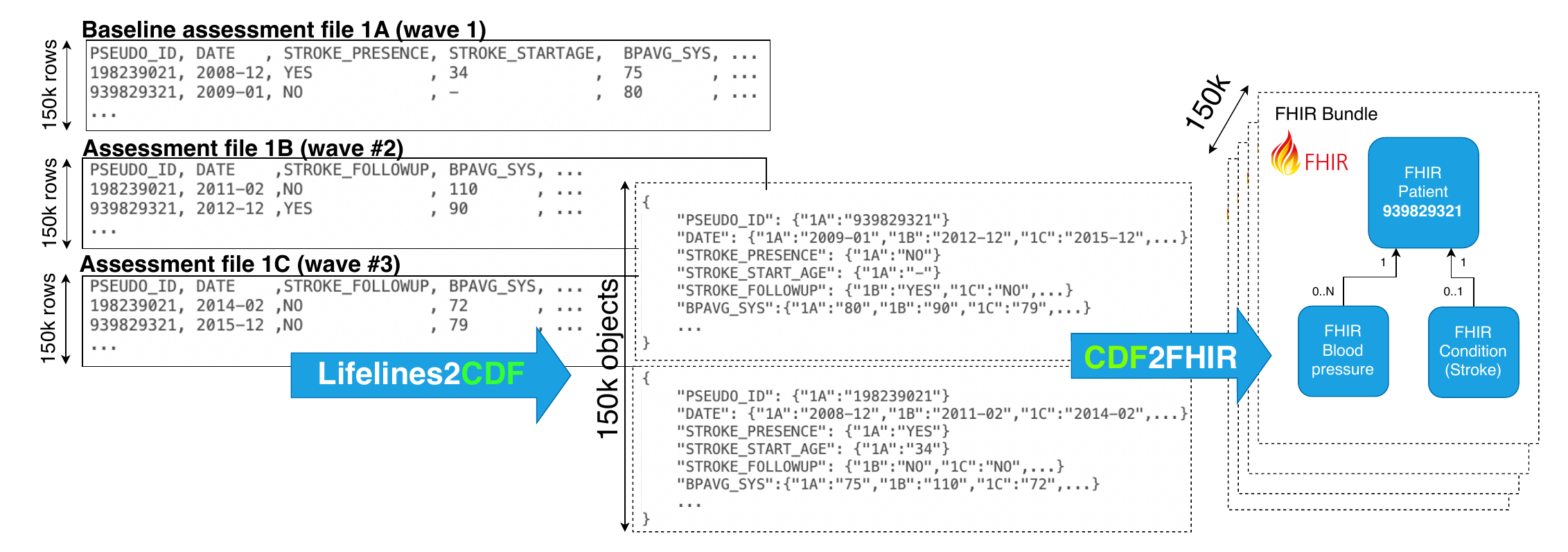}
	\caption{Left: Lifelines datafile format, where the assessments of each data collection phase (a wave) are available on a separate file. Center: Intermediate representation format used by the general-purpose data-harmonization tools. Right: Generated FHIR resources. Lifelines2CDF: tool for transforming Lifelines data files into the CDF format. CDF2FHIR: tool for transforming CDF records into FHIR resources, given a set of \textit{pairing rules}. }
	\label{fig:intermediate-format}
\end{figure}

To illustrate the limited reusability of a dataset-specific harmonization pipeline, consider the process of transforming \textit{Lifelines} data into FHIR-compliant resources. As depicted in the left part of Fig.~\ref{fig:intermediate-format}, in Lifelines the phenotype data of each data collection phase (i.e., `waves' 1A, 1B, 1C, 2A, etc.) are scattered across multiple files. Consequently, interpreting pairing rules such as the one listed in Algorithm~\ref{alg:stroke_onset} would require a very specific process to efficiently manage the repeated (and prohibitively expensive) access to multiple data files for each pairing rule and each of the nearly 150,000 Lifelines participants. Consequently, any tool developed specifically for Lifelines would not be transferable to other datasets with different structures, forcing the creation of new ones every time. This lack of a unified approach, in addition to the inefficiencies it will carry in the long run, will cause maintainability issues, as optimizations and updates made for one tool would not be transferable to others.

\begin{algorithm} \caption{Pairing rule that maps Lifelines data with the \textit{start date} of a  ZIB-FHIR's Stroke Problem~\cite{zibsProblemv412017ENZorginformatiebouwstenen}}\label{alg:stroke_onset} \begin{algorithmic}[1]
		
		\Function{approximateStrokeOnsetDate}		
		
		\If{value("STROKE\_PRESENCE", "1A") $=$ 'yes'}:
		
		\State surveyYear $\gets$ year(value("DATE", "1A"))
		\State firstAssessmentAge $\gets$ value("AGE", "1A")
		\State strokeStartAge $\gets$ value("STROKE\_STARTAGE","1A")
		
		\If{strokeStartAge $\not=$ \textit{undefined} \textbf{and} firstAssessmentAge $\not=$ \textit{undefined}}:
		\State \Return (surveyYear - firstAssessmentAge + strokeStartAge)
		\Else:
		\State \Return \textit{undefined}
		\EndIf
		
		\Else:
		\State strokeFollowUpValues $\gets$ values("STROKE\_FOLLOWUP",["1A","2B","1C","2A","3A","3B"])
		\State strokeFollowUpDates $\gets$ values("DATE",["1A","2B","1C","2A","3A","3B"])
		\State timeInterval $\gets$ strokeReportInterval(strokeFollowUpValues,strokeFollowUpDates)
		\If{timeInterval $\not=$ \textit{undefined}}:
		\State previousNoStrokeAssessmentDate, strokeAssessmentDate $\gets$ timeInterval
		\State \Return dateToISO(meanDate(previousNoStrokeAssessmentDate, strokeAssessmentDate))
		\Else:
		\State \Return undefined
		\EndIf
		\EndIf
		\EndFunction
		
\end{algorithmic} \end{algorithm}

To make this process scalable within the MyDigiTwin research infrastructure, a generic reusable harmonization framework~\cite{Cadavid_CDF2FHIR,cadavid2024leveraging}, applicable to any longitudinal observational study data (where multiple phenotype variables are collected over time) was developed. In line with the adoption of the ZIB-FHIR profile for consistency between model's training data and the prediction's input data, the design of this reusable data-harmonization framework was focused on reliability (i.e., transforming data consistently and accurately) and robustness (i.e., handling unexpected or irregular data inputs) qualities. This is important considering that the complexity of the target FHIR profile, the complexity of pairing rules (such as that of the Algorithm~\ref{alg:stroke_onset}), and the human elements involved, makes the overall process error-prone. 

\begin{figure}[th]
	\centering
	\includegraphics[width=0.7\linewidth]{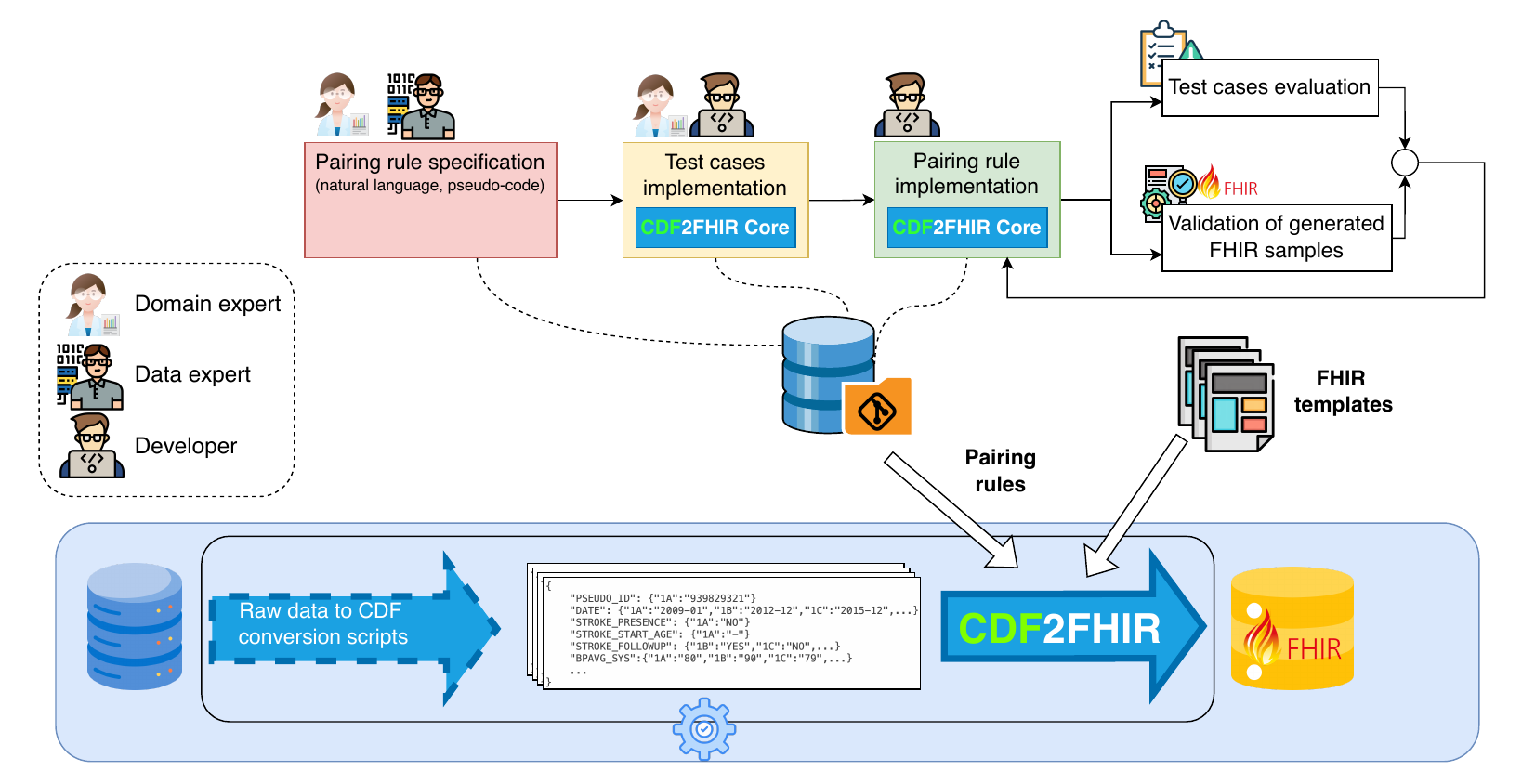}
	\caption{\textit{Pairing rules} development workflow. Top: the git-based collaborative environment for developing pairing rules following a test-driven approach. Bottom: the pairing rules, once defined, are used on a batch transformation process --creating the target FHIR dataset-- running where the data resides.}
	\label{fig:diagrams-devflow}
\end{figure}

\begin{figure}[th]
	\centering
	\includegraphics[width=0.7\linewidth]{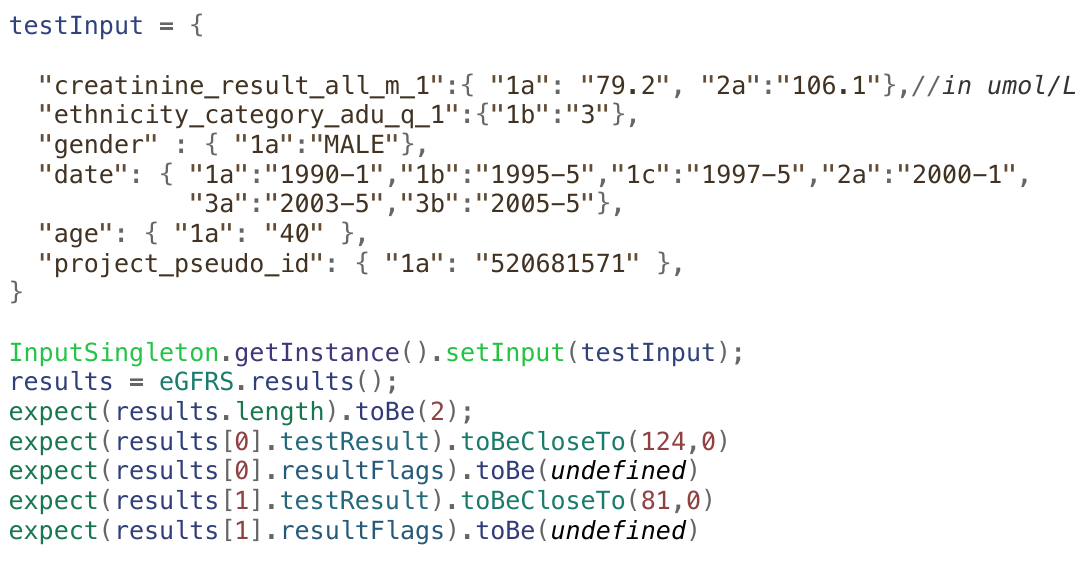}
	\caption{One of the test cases for the mapping \textit{pairing rules }between Lifelines data and the ZIB's Laboratory Test Result for Glomerular Filtration Rate (GFR), which are expected to work according to the 2009 CKD-EPI creatinine and creatinine-cystatin C equations~\cite{inker2012estimating}.}\label{fig:sample_unit_test}
\end{figure}

One of the key features of this framework that improves the overall robustness of the harmonization process, is that it supports the development of \textit{pairing rules} with the rigor and structure of quality software engineering. This involves enabling a test-driven development (TDD) workflow~\cite{beck2022test} for such rules, integrated within a Git version control system, to ensure iterative improvement and traceability of changes. In this workflow, as depicted in Fig.~\ref{fig:diagrams-devflow}, medical domain experts and data experts cooperate through platforms like GitHub or GitLab in the high-level specification of the \textit{pairing rules} using natural language and pseudocode. Based on this specification, medical domain experts and developers work collaboratively on the definition of test cases that will guide the implementation of the pairing rules (see Fig.~\ref{fig:sample_unit_test}).

Upon successful implementation --achieved when all test cases pass-- an additional validation step is performed to ensure conformance of FHIR resources generated from ``dummy'' test data. This final step verifies that the pairing rules, when used to provide values while generating FHIR resources, meet requirements such as accepted codes, cardinality rules, data types, and other specifications from the ZIB profile. When all these quality checks have been passed, the pairing rules can then be executed as a batch process, running within the premises where the data resides, to perform the transformation of the actual data set (see the bottom part of Fig.~\ref{fig:diagrams-devflow}) into a FHIR dataset, and then integrated --through a flattening process~\cite{Cadavid_SQLonFHIRProjections}-- into the federated learning architecture previously described in Fig.~\ref{fig:fhir-for-training-prediction-consistency}.

Making this tool generic and reusable, however, requires decoupling it from the data management strategy of the source raw data. To this end, an intermediate \textit{ cohort data format} (CDF) was defined as its standard input format. As illustrated in Fig.~\ref{fig:intermediate-format}, this intermediate format not only preserves the details of the original, but also arranges it in a way that allows \textit{pairing rules}, such as the one in Algorithm~\ref{alg:stroke_onset}, to directly access any variable from any data collection phase of a given patient, making the overall process more efficient in terms of temporal complexity. While this requires an additional step of restructuring the raw data of each new reference dataset into the CDF format, it is significantly simpler —by an order of magnitude— than building an entirely new tool to convert raw data directly into FHIR. Creating such a tool would involve developing complex artifacts from scratch, including mechanisms for interpreting pairing rules and ensuring the consistent generation of FHIR output.

The following section elaborates on how the research infrastructure and related tools previously described were evaluated through an end-to-end proof of concept. That is, starting from the problem of harmonizing a big-data reference dataset like \textit{Lifelines}, up to the development of a prediction model trained using the federated learning infrastructure.

\section{Proof of concept}\label{sec:poc}

This section outlines a proof-of-concept of the MyDigiTwin framework, aiming to validate that the proposed data harmonization pipeline and the federated research infrastructure, together, can effectively support the development of 10-year cardiovascular disease (CVD) risk prediction models that could serve the purpose of performing predictions within a \textit{personal health environment}, as described in Section~\ref{sec:architecture}. In clinical practice, risk prediction scores play a crucial role in guiding CVD treatment initiation and intensification. The UK National Institute for Health and Care Excellence recommends using the QRISK3 \cite{hippisley_cox_development_2017_qrisk3} tool to estimate 10-year cardiovascular disease risk. The American College of Cardiology and the American Heart Association advocates for the Atherosclerotic Cardiovascular Disease (ASCVD) risk score \cite{arnett_2019}, while the European Society of Cardiology advocates the SCORE2 \cite{score2_working_group_and_esc_cardiovascular_risk_collaboration_score2_2021} model for risk prediction in Europe. These algorithms help tailor treatment strategies based on individual risk profiles.

The first aspect validated in this proof-of-concept was that a big data reference data set can be consistently transformed into a ZIB-profile-compliant FHIR dataset. This analysis used the Lifelines dataset and focused on 12 commonly used predictors for cardiovascular disease, utilized in widely available CVD prediction scores. These predictors included age, sex, kidney markers (e.g., eGFR, albumin), biomarkers (e.g. HDL cholesterol, HbA1c), disease histories (e.g., hypertension, type 2 diabetes), and lifestyle factors (e.g., smoking status, BMI).  This selection reflects variables with strong and well-documented associations with CVD risk~\cite{arnett_2019, benjamin_heart_2019, noauthor_preventtm_2024,dziopa_cardiovascular_2022}, ensuring that the implementation of the proof of concept aligns with clinically validated predictors. In addition to these predictors, four variables were included in the harmonization process as the outcomes of the CVD risk prediction, namely stroke, myocardial infarction (MI), heart failure (HF), and a composite CVD outcome (stroke, MI, HF); the outcome definitions are based on the composite endpoint of those, summarized in \Cref{tab:lifelinesvalidationvars}, during the 10-year follow-up period. The pairing rules for both predictors and outcomes were specified through a collaborative process following the workflow described in Section~\ref{sec:harmonization}, involving the Lifelines data manager and health domain specialists with expertise in the FHIR specification. The specifications for these pairing rules, along with their corresponding test cases and the discussions that shaped the development of their current versions, are openly available within the project's GitHub organization\footnote{\url{https://github.com/MyDigiTwinNL/CDF2Medmij-Mapping-tool/tree/main/src/lifelines}}.

After implementing and running the transformation batch process on Lifelines' premises, we validated the consistency of the FHIR dataset in two independent experiments. As depicted on parts one and two of Fig.~\ref{fig:poc-diagram}, we performed a data analysis on the flattened version of the harmonized FHIR data using SQL queries. For verification, we conducted an equivalent analysis directly on the Lifelines ``raw'' variables listed in table~\ref{tab:lifelinesvalidationvars} using a series of independently implemented scripts. The validation was carried out by comparing the results of both approaches to ensure consistency and accuracy in the harmonized data set.

\begin{table}[th]
  \caption{Outcome definitions using Lifelines variables.}
  \label{tab:lifelinesvalidationvars}
  \includegraphics[width=0.7\linewidth]{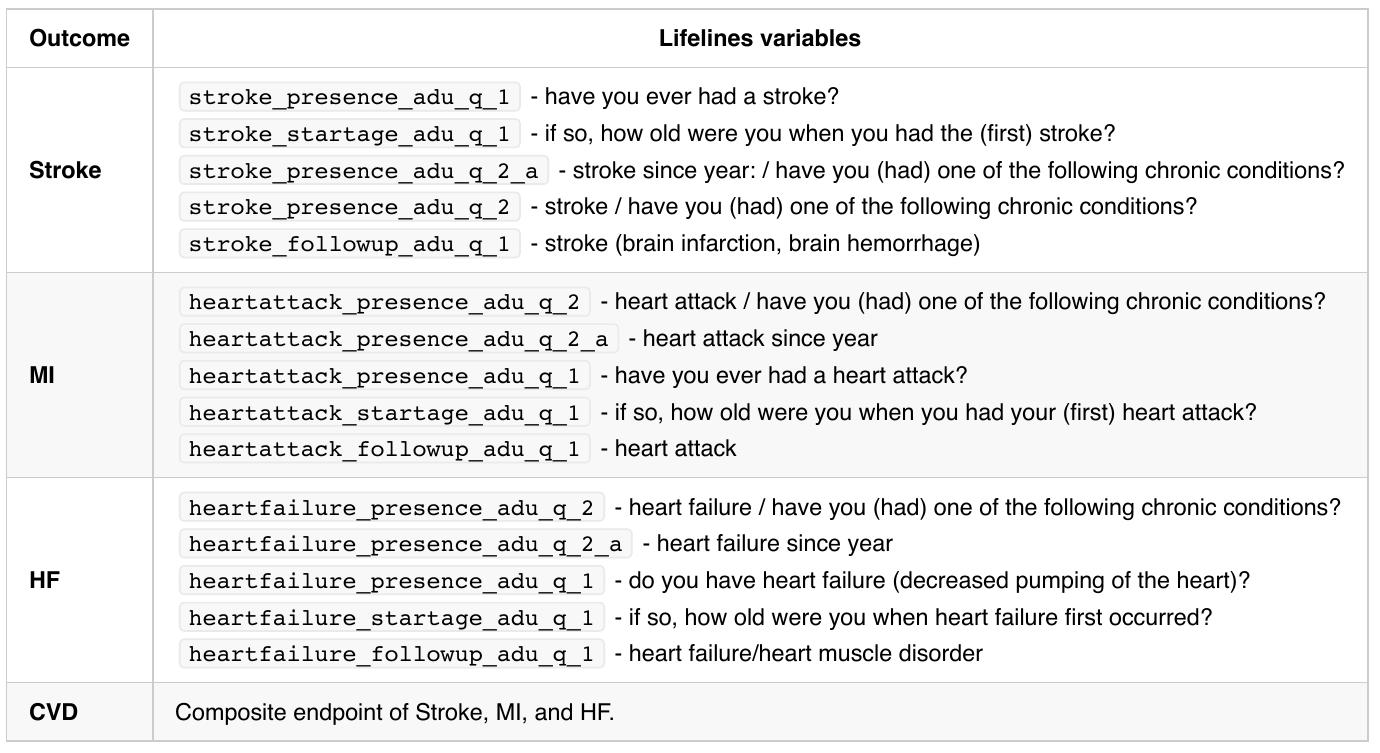}
\end{table}

\subsection{Research infrastructure setup}\label{ssec:res_infra}

To evaluate the federated learning process within the Lifelines cohort, which requires multiple data stations, a simulated federated environment was created by splitting the harmonized Lifelines dataset into three subsets, each assigned to a separate vantage6 node. Although this setup does not fully replicate key aspects of a real federated scenario—such as statistical heterogeneity across institutions—it effectively serves to test the core functionalities of federated learning, including data distribution, local model updates, and aggregation processes.

\begin{figure}
	\centering
	\includegraphics[width=0.7\linewidth]{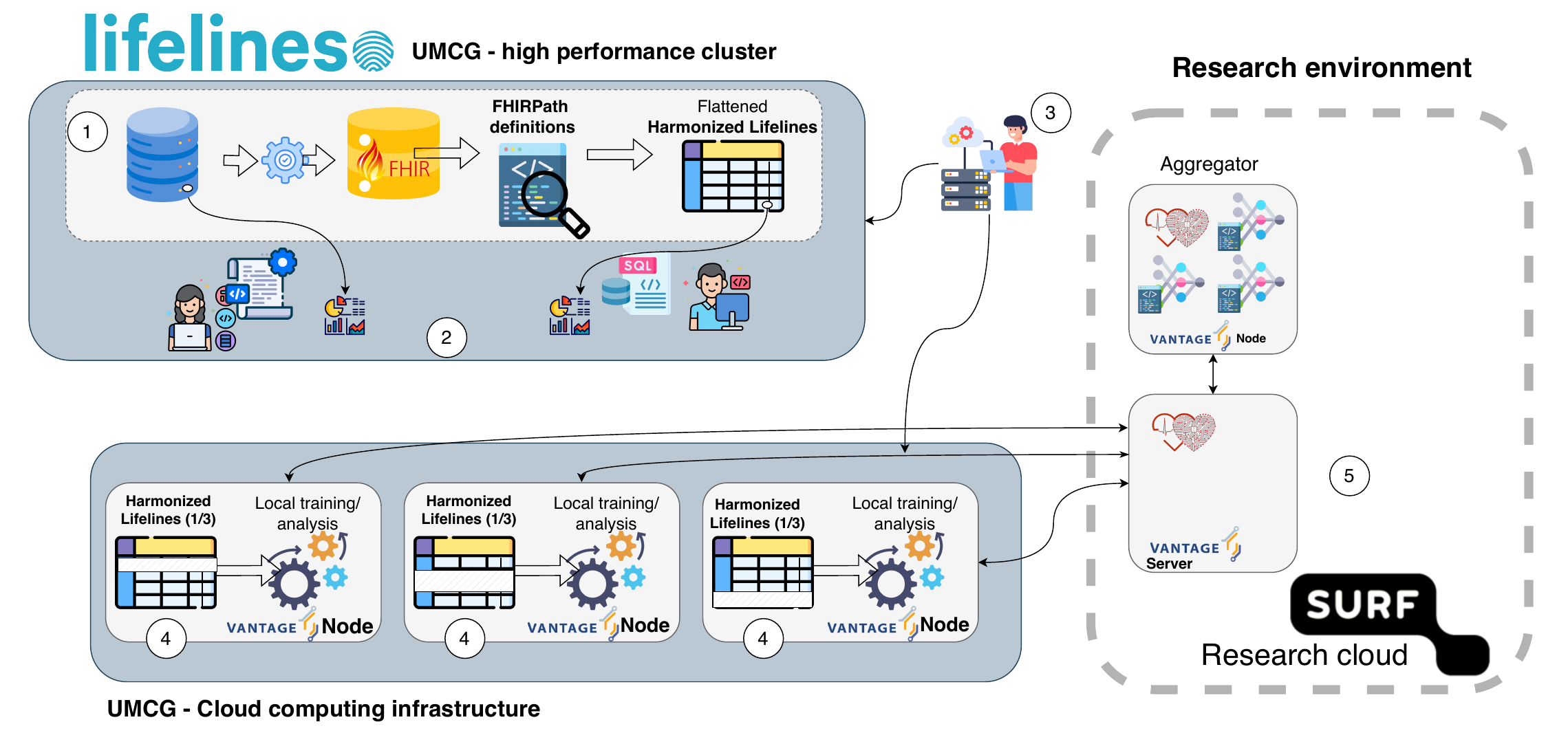}
	\caption{Research infrastructure setup for the MyDigiTwin's proof of concept: (1) Data harmonization pipeline running on Lifelines HPC infrastructure, (2) Local data analyses performed on both raw and harmonized data for validation purposes. (3) Lifelines data manager moving the flattened version of the harmonized data to the vantage6 nodes, (4) simulation of two nodes by splitting the datasets, (5) server and aggregator node running on SURF infrastructure.}
	\label{fig:poc-diagram}
\end{figure}

As shown in Fig.~\ref{fig:poc-diagram}, the Lifelines data manager played a central role in managing the infrastructure. With access to both the HPC infrastructure (used for data harmonization) and the cloud environment (hosting the vantage6 nodes), the data manager was responsible for transferring and updating harmonized datasets between these environments. Additionally, the data manager established security constraints, such as defining permissible algorithms and specifying the datasets accessible for analysis. These operational and node-management actions have been documented as guidelines to support future organizations joining the federated infrastructure as data stations\footnote{\url{https://github.com/MyDigiTwinNL/MyDigiTwin-federated-learning-node-setup-guidelines}}.

The vantage6 server and an additional vantage6 aggregator node were configured within the SURF Research Cloud infrastructure to ensure that the aggregation of data and results was performed in a neutral and trusted environment. This setup guarantees both the privacy of distributed datasets and the reliability of aggregated model outputs.

\subsection{Model development}\label{ssec:development}
The proof-of-concept comprises the development of a federated learning algorithm for model training and its validation on harmonized (FHIR) data in different client nodes, as illustrated in the bottom right of \Cref{fig:fhir-for-training-prediction-consistency}. The baseline prediction model is a neural network. Our experimentation aims to verify that the performance of the prediction model trained using federated learning on horizontally partitioned data is higher than that of a prediction model trained on a single node. In our experiments, Harrell's C-statistic \cite{harrell1982evaluating} evaluates the discrimination performance of the prediction model. 

\subsubsection{Federated learning algorithm}\label{sssec:fl}
To apply a federated learning algorithm, as explained in \Cref{sec:architecture}, we consider harmonizing a big-data reference dataset through FHIR data that follow a set of FHIRPath expressions. Based on the defined FHIRPath expressions, the flattened FHIR data (i.e., two-dimensional tabular data) in different nodes always have the same feature space (i.e., columns in the tabular format) comprising the selected predictors and the defined outcomes in this section. The harmonized data in different nodes can be considered as horizontally partitioned inside a tabular view of a database. Here, horizontal partitioning is to split rows of a table into different groups; each row of the partitions has the same and complete data features. We can then employ sample-partitioned federated learning, so-called horizontal federated learning (HFL) \cite{kairouz2021advances}, on them. \par
As outlined in \Cref{sec:architecture}, the MyDigiTwin Framework has a client-server architecture \cite{yang2019federated} of a horizontal federated learning system. Here, a client indicates a node conducting local training of a prediction model on the local data (horizontally partitioned data), and a server coordinates an aggregator node that aggregates the local models trained on each independent client node. In particular, the baseline aggregation approach used in our PoC is model averaging \cite{mcmahan2017communication}, which uses an average of locally trained models' weight as the aggregated model weight. In the rest of the paper, we denote this federated learning via model averaging as federated averaging (FedAvg)\footnote{strictly speaking, \textit{federated averaging} involves both gradient averaging and model averaging. However, in the literature (as well as in the rest of this paper), \textit{federated averaging} is generically used to refer to the used aggregation approach, i.e., \textit{model averaging}.} \par
\begin{figure}[th]
	\centering
	\includegraphics[width=0.6\linewidth]{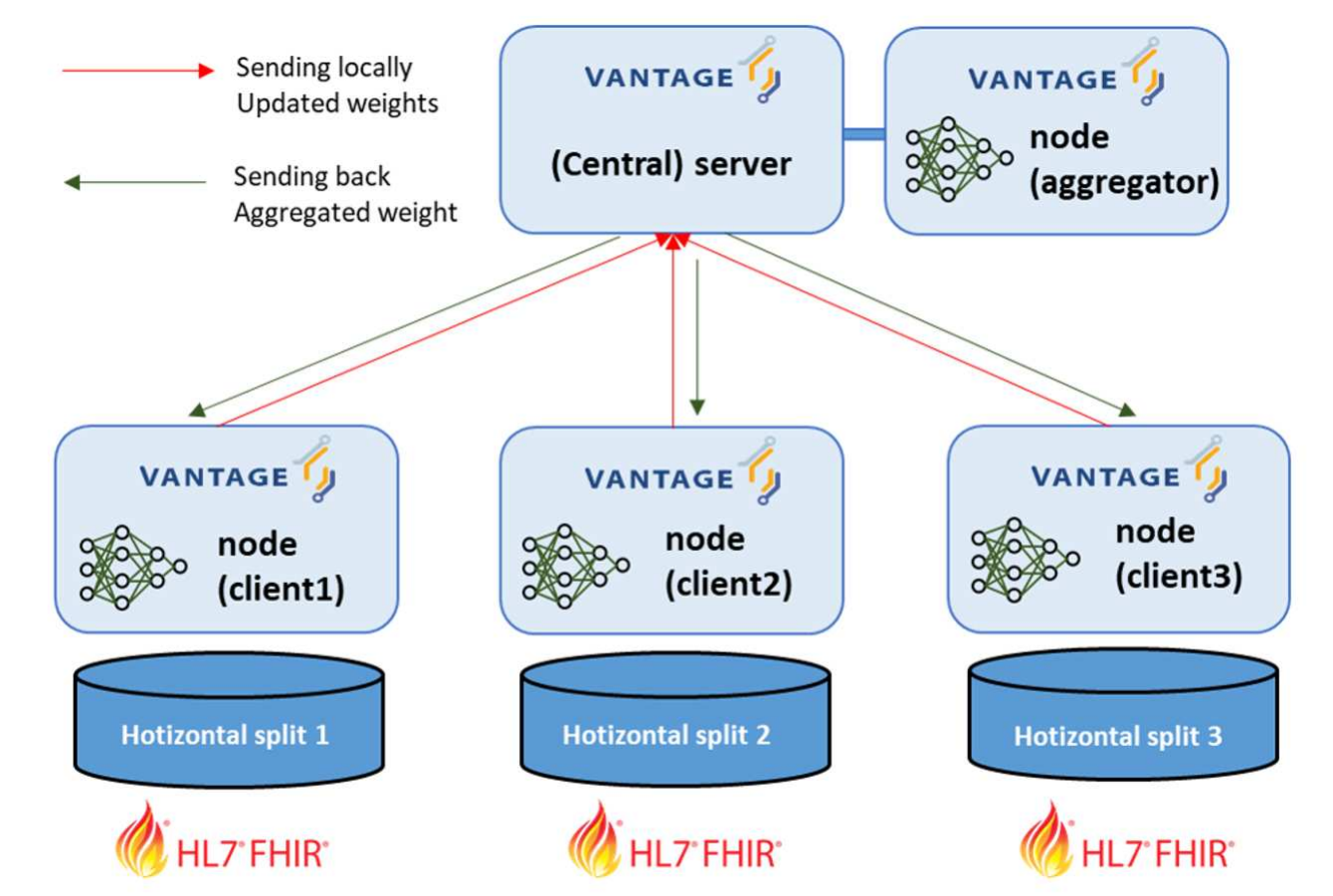}
	\caption{Overview of federated learning with vantage6 infrastructure, on horizontal splitter data.}
	\label{fig:fl_arch_overview}
\end{figure}

The pseudo-code of the FedAvg algorithm adapted in our work is shown in \Cref{alg:fedavg}. As depicted in \Cref{fig:fl_arch_overview}, all client nodes contain a neural network (our baseline prediction model explained in \Cref{sssec:models}) with the same architecture. At the beginning of the algorithm, the aggregator node initializes model weight $w_{0}$, and the server broadcasts it to all $K$ client nodes ($K$=3 in our PoC). An iteration of the aggregation starts with local training at each client node on the data they have. Here, the local weight update method is gradient descent (GD), one of the most commonly used optimization algorithms for deep learning (DL). Namely, for each client $k$, the broadcasted weight $\overline{w}_{t}$ is updated to $w^{k}_{t+1}$. It is then sent back to the server. When the server receives $w^{k}_{t+1}$ from all $K$ clients, the aggregation node computes the weighted average of them as specified in line \ref{line:averaging} of \Cref{alg:fedavg} where $n_{k}$ is the number of training samples in node $k$. Finally, the server broadcasts the aggregated model weight $\overline{w}_{t+1}$ to all the clients for the next iteration in the loop.  \par
\algnewcommand\algorithmicinput{\textbf{Initialization by the server:}}
\algnewcommand\INIT{\item[\algorithmicinput]}
\begin{algorithm}
\caption{The FedAvg algorithm}\label{alg:fedavg}
\begin{algorithmic}[1]
\State \textbf{Initialization by the server:}
\Indent 
\State The aggregator node initializes model weight $w_{0}$, and the server broadcasts it to all client nodes. 
\EndIndent
\newline 
\For{each global model update iteration $i \in \{1,2,..., I\}$}
\For{each client node $k \in \{1,2,..., K\}$ \textbf{in parallel}}
\State Local training on the local data: $w^{k}_{i+1} \gets \textbf{Gradient descent optimization}(k, \overline{w}_{t})$ \label{line:local_update} 
\State Sending back $w^{k}_{i+1}$ to the server. \label{line:return} 
\Comment{$w^{k}$: local model weight, $\overline{w}$: aggregated (averaged) weight}
\EndFor
\State The server receives local model weights, $w^{k}_{i+1}$ for all $k$.
\State The aggregation node takes the weighted average: $\overline{w}_{t+1} \gets \sum^{K}_{k=1}\frac{n_{k}}{n}w^{k}_{t+1}$ \label{line:averaging}
\Comment{$n = \sum^{K}_{k=1}n_{k}$}
\State The server broadcasts $\overline{w}_{t+1}$ to all the clients. \label{line:aggr_broadcast}
\EndFor
\end{algorithmic}
\end{algorithm}

\subsubsection{Prediction models}\label{sssec:models}
The prediction model used in our PoC is a deep learning implementation of the Cox proportional hazards (PH) regression models \cite{cox1972regression}. The log-risk function of the Cox PH model is a linear combination of predictors, while our method estimates it using a feedforward neural network that takes predictors as input. I.e., the network outputs the log-risk function for the given predictors. This idea of deep learning implementation of Cox models, described above, was proposed by Katzman et al. \cite{katzman2018deepsurv}. \par
The architecture of the neural network used in our experiments comprises four fully connected layers. The input layer has $p$ neurons, where $p$ denotes the number of considered predictors. It is followed by the two hidden layers that have 32 neurons each. The output layer is a single linear neuron. Thus, the architecture can be denoted in this compact notation: [$p$, 32, 32, 1]. ReLU is used as an activation function throughout the entire network. A dropout of 0.25 is applied between each layer. \par 

\subsubsection{Datasets}\label{sssec:datasets}
In the experiments, we used two datasets: Worcester Heart Attack Study (WHAS) and  Lifelines. Common to both datasets, we horizontally partitioned the dataset into three subsets having 50$\%$, 30$\%$, and 20$\%$. We then put one subset on each client. In each client, we further split the data in a ratio of 80:10:10 for training, validation, and test. The WHAS was used to implement the prediction model and its federated learning with vantage6. We used the Lifelines to validate the improvement of predictive accuracy in CVD risk predictions using the implemented model and its federated learning.  \par

\paragraph{Worcester Heart Attack Study (WHAS) Dataset}
The Worcester Heart Attack Study (WHAS) dataset \cite{hosmer2008applied} is a public dataset that consists of 1,638 subjects from WHAS \cite{goldberg2000decade} that intended to investigate factors and temporal trends associated with long-term survival following acute myocardial infarction (MI) among inhabitants of Worcester, Massachusetts, United States. It comprises 5 predictors: age, sex, body mass index (BMI), congestive heart failure (CHF), and order of MI. The CHF predictor indicates the absence or presence of congestive heart failure in patients. The MI order presents whether the acute myocardial infarction of a patient is initial or prior. The outcome definition is survival as of the last follow-up. It should be noted that this dataset is mainly used for the implementation of the regression model and its evaluation verifying whether the aggregated model by the FL improves predictive performance, although the model is not intended to predict CVD risk for individuals who do not have prevalent CVD. We did not harmonize the WHAS data into FHIR format since 
the harmonization does not affect the implementation and performance evaluation mentioned above. In our experimental setup, out of the given 1,638 subjects, 819 are located in the first client node (491 / 164 / 164 for training / validation / test); the second node has 491 (295 / 98 / 98); the remaining 328 are placed in the last node (197 / 65 / 65). \par

\begin{table}[hb!]
 \caption{Participant characteristics in the Lifelines dataset used in our PoC.}
 \label{tab:characteristics_liflines}
 \centering
 \begin{tabular}{lcc}
 \toprule
 Clinical characteristic &  Mean (SD) or N(\%) & Median (Q1; Q3)  \\
 \midrule
Total number of individuals & $148,230$ &   \\
Follow-up, years &  & $6.5$ $(2.6; 10.0)$  \\
Women &  $87,381$ $(58.9\%)$  &\\
Age, years &  $44.4$ $(13.1)$  & $44.2$ $(35.4; 51.3)$\\
Systolic blood pressure, mmHg & $127.1$ $(16.1)$ & $125$ $(116; 137)$  \\
Diastolic blood pressure, mmHg & $73.7$ $(9.5)$ & $73$ $(67; 80)$   \\
HDL-cholesterol, mmol/L & $1.5$ $(0.4)$ & $1.5$ $(1.2; 1.7)$   \\
LDL-cholesterol, mmol/L & $3.3$ $(0.9)$ & $3.2$ $(2.6; 3.9)$   \\
Total-cholesterol, mmol/L & $5.1$ $(1.0)$ & $5.0$ $(4.4; 5.7)$   \\
EGFR, ml/min/1.73 \text{$m^2$} & $91.2$ $(16.0)$ & $91.6$ $(80.2; 102.7)$   \\
Albumin, g/L & $45.1$ $(2.4)$ & $45$ $(43; 47)$   \\
HbA1c, mmol/mol & $36.7$ $(5.1)$ & $36$ $(34; 69)$   \\
Creatinine, umol/L & $76.2$ $(14.7)$ & $75$ $(66; 85)$   \\
Smoking quantity & $12.1$ $(11.2)$ & $9$ $(3.8; 17.3)$   \\
Smoking status & $$ & $$   \\
\quad Never smoked & $65,335$ $(45.8\%)$ &   \\
\quad Ex-smoker & $51,986$ $(36.5\%)$ &    \\
\quad Current smoker & $25,273$ $(17.7\%)$&   \\
Type 2 diabetes & $3,998$ $(2.7\%)$ &   \\
Hypertension & $36,557$ $(24.7\%)$ &    \\
 \bottomrule
 \end{tabular}
\end{table}

\paragraph{Lifelines}
The Lifelines dataset used in our experiments is the harmonized data of the LifeLines Cohort Study with the target FHIR profile specified in \Cref{sec:harmonization}. The LifeLines Cohort Study, established in 2006, is a large population-based prospective study designed to enhance our understanding of healthy aging among residents of the Northern part of the Netherlands. \par
We included the predictors outlined in \Cref{sec:poc}. In particular, the used predictors consist of the following 15 variables: age, sex, eGFR, albumin, HDL cholesterol, LDL cholesterol, total cholesterol, HbA1c, hypertension history, type 2 diabetes history, creatinine, systolic blood pressure, diastolic blood pressure, smoking history, smoking quantity. The outcome measure is incident CVD composed of stroke, MI, and HF (see also \Cref{sec:poc}).
The harmonized Lifelines dataset used in our experiments contains 151,078 subjects. One subject without a birth date is excluded. We further excluded 2,847 subjects with prevalent CVD. The baseline characteristics of the individuals are summarized in \Cref{tab:characteristics_liflines}. In our experimental setup, from those 148,230 subjects, 74,115 are located in the first client node (44,469 / 14,823 / 14,823 for training / validation / test); the second node has 44,469 (26,681 / 8,894 / 8,894); the remaining 29,646 are placed in the last node (17,787 / 5,929 / 5,930). \par

\begin{table}[ht!]
\footnotesize
\begin{center}
\caption{Results of the model performance evaluation on the 1,638 subjects of the WHAS dataset, in terms of C-statistic with 95$\%$ CIs through a corrected resample t-test based on 10 independent runs.}\label{tab:whas_results}
\begin{tabular}{l>{\centering}p{4cm}>{\centering\arraybackslash}p{4cm}}
\toprule[1pt] 
\multirow{2}*{Model evaluation}& \multicolumn{2}{c}{C statistic (95\%CI)}\\
& Without aggregation & FedAvg (after 20 updates) \\
\midrule[0.5pt]
Global & 0.755 (0.737-0.774) & 0.775 (0.760-0.790)  \\
Local (on client1) & 0.773 (0.739-0.806) & 0.793 (0.753-0.832)   \\
Local (on client2) & 0.777 (0.738-0.816) & 0.791 (0.753-0.829)   \\
Local (on client3) & 0.712 (0.643-0.780) & 0.725 (0.638-0.811)   \\
\arrayrulecolor{black}\bottomrule[1pt]
\noalign{\vskip 0.3mm} 
\end{tabular}
\end{center}
\end{table}

\begin{figure}[t!] 
 \centering
 \subfloat[\label{fig:fl_glob_results_whas}]{%
 \includegraphics[width=0.44\linewidth]{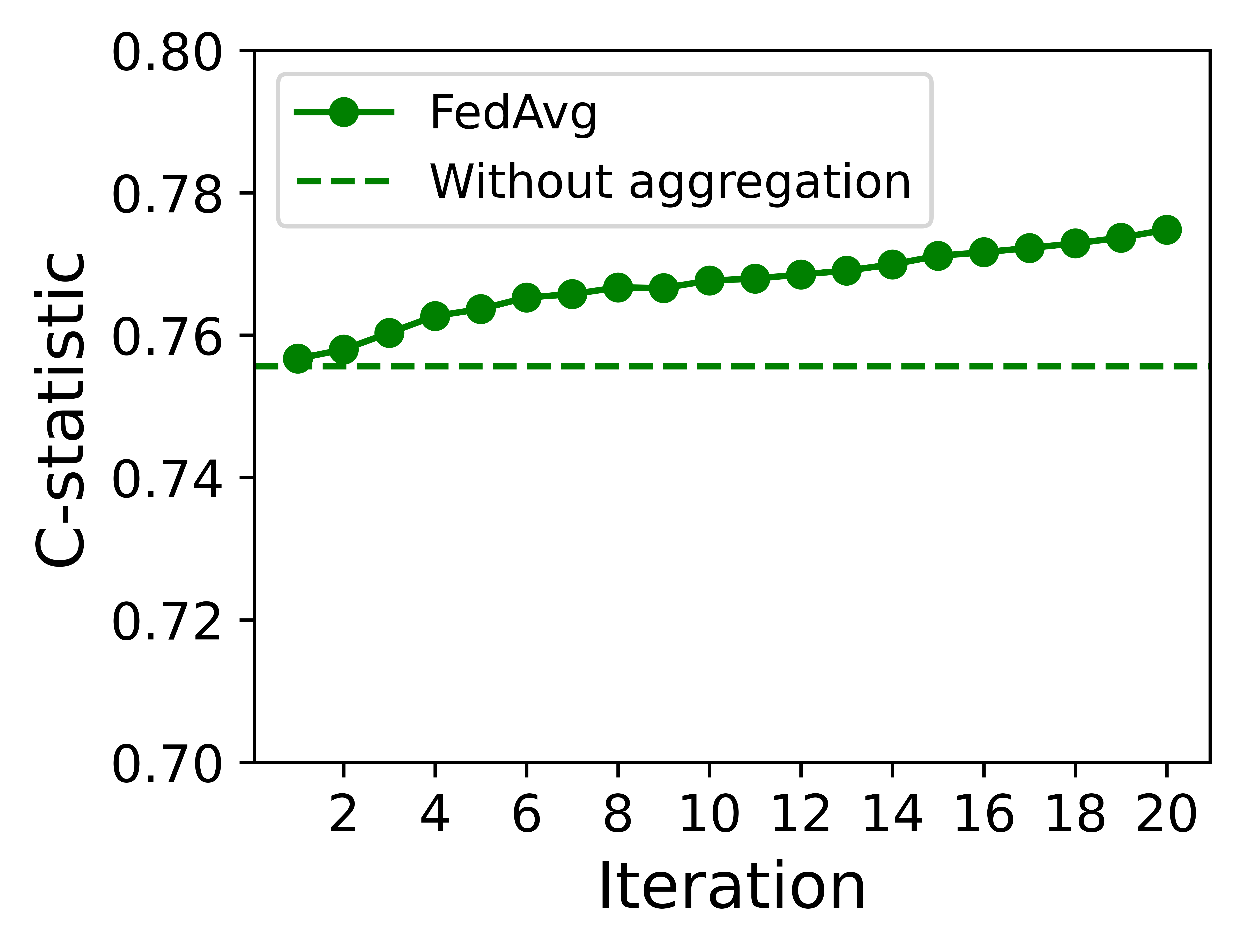}}
 \qquad
 \subfloat[\label{fig:fl_local_results_whas}]{%
 \includegraphics[width=0.44\linewidth]{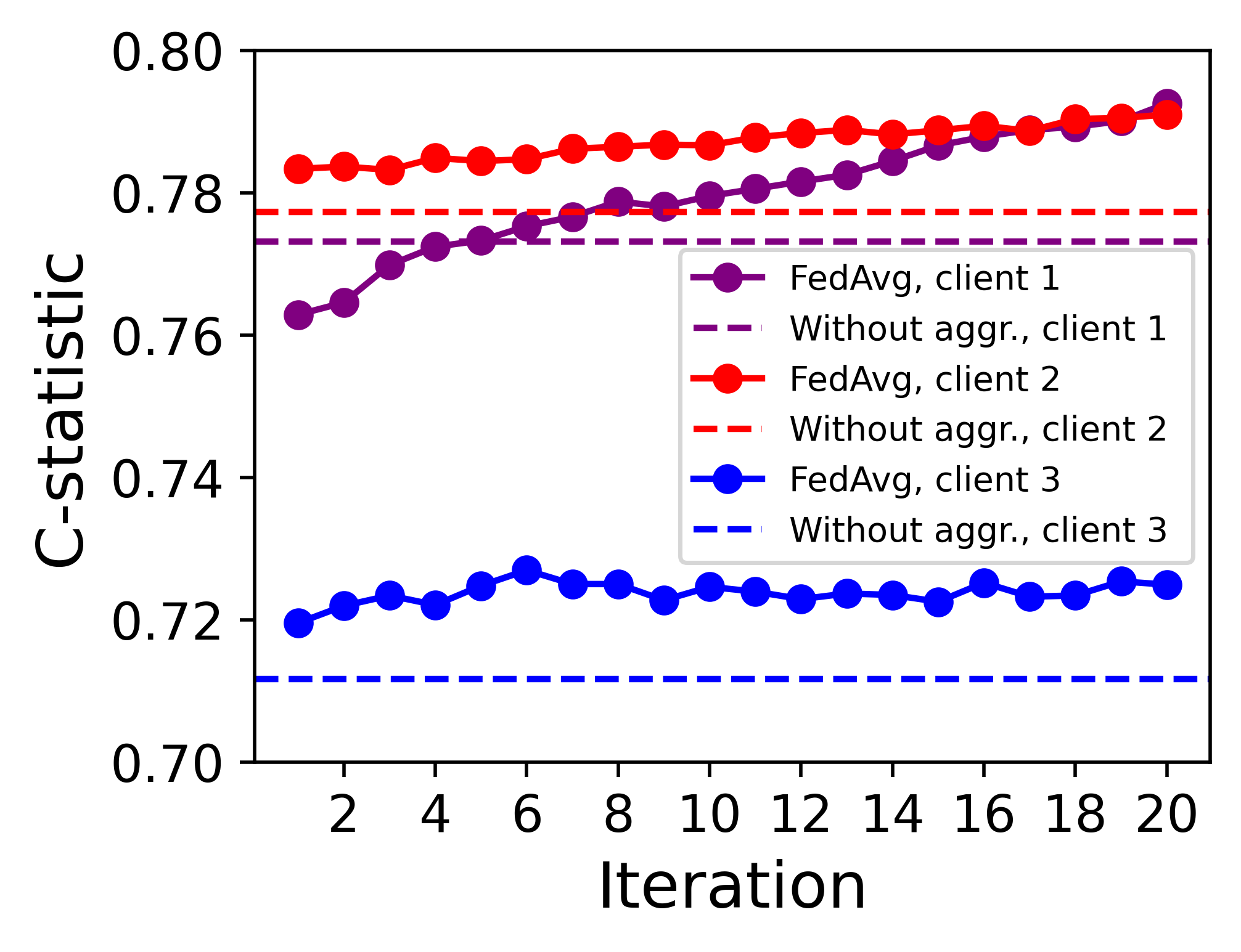}} 
 \caption{C-statistic curve presenting the discriminative ability of the proposed method, \emph{FedAvg}, compared to \emph{Without aggregation} on the WHAS dataset (a) the global model evaluation (b) the local model evaluation.}
 \label{fig:fl_results_whas} 
\end{figure}


\subsection{Validation}\label{ssec:validation}
We implemented the prediction model with federated learning using the WHAS dataset and evaluated its performance. An improvement in predictive performance by the proposed method was validated on the Lifelines dataset. Our experiments aim to confirm a gradual improvement in the model's predictive performance across the iterations of the aggregation for all test data in different clients. Therefore, we focus on monitoring the discriminative performance of the model in C-statistic during update iterations of federated learning. Since this is a PoC, only the discriminative ability is included in this work but it should be noted that evaluating calibration metrics could be considered an additional analysis before applications of the proposed method. The current implementations are publicly available in the project's GitHub organization\footnote{\url{https://github.com/MyDigiTwinNL/FedAvg_vantage6}}.
\par
All partitioned data of the datasets \Cref{sssec:datasets} are located in the client nodes and always remain secure within their original location. The missing predictor values were separately handled in each client using multiple imputation by chained equations. The predictor values were then rescaled into the range $[0,1]$ for our prediction model (a neural network), using min-max normalization for each client independently. \par
As illustrated in \Cref{fig:fl_arch_overview}, model training is carried out at each client independently; the local training data are used for the network weights optimization, and validation loss to monitor overfitting is measured using the local validation data. After this model training (line \ref{line:local_update} of \Cref{alg:fedavg}), the model makes predictions on the local test data, and we evaluated the local model performance of a single client node. Those predictions are also collected to evaluate the global model performance; we collected all the prediction results on the local test data from all the client nodes and computed the C-statistic. Thus, the local and global model performance was evaluated by the prediction results on the local test data that has not been used for training. \par
\begin{figure}[t!] 
 \centering
 \subfloat[\label{fig:fl_glob_results_lifelines}]{%
 \includegraphics[width=0.44\linewidth]{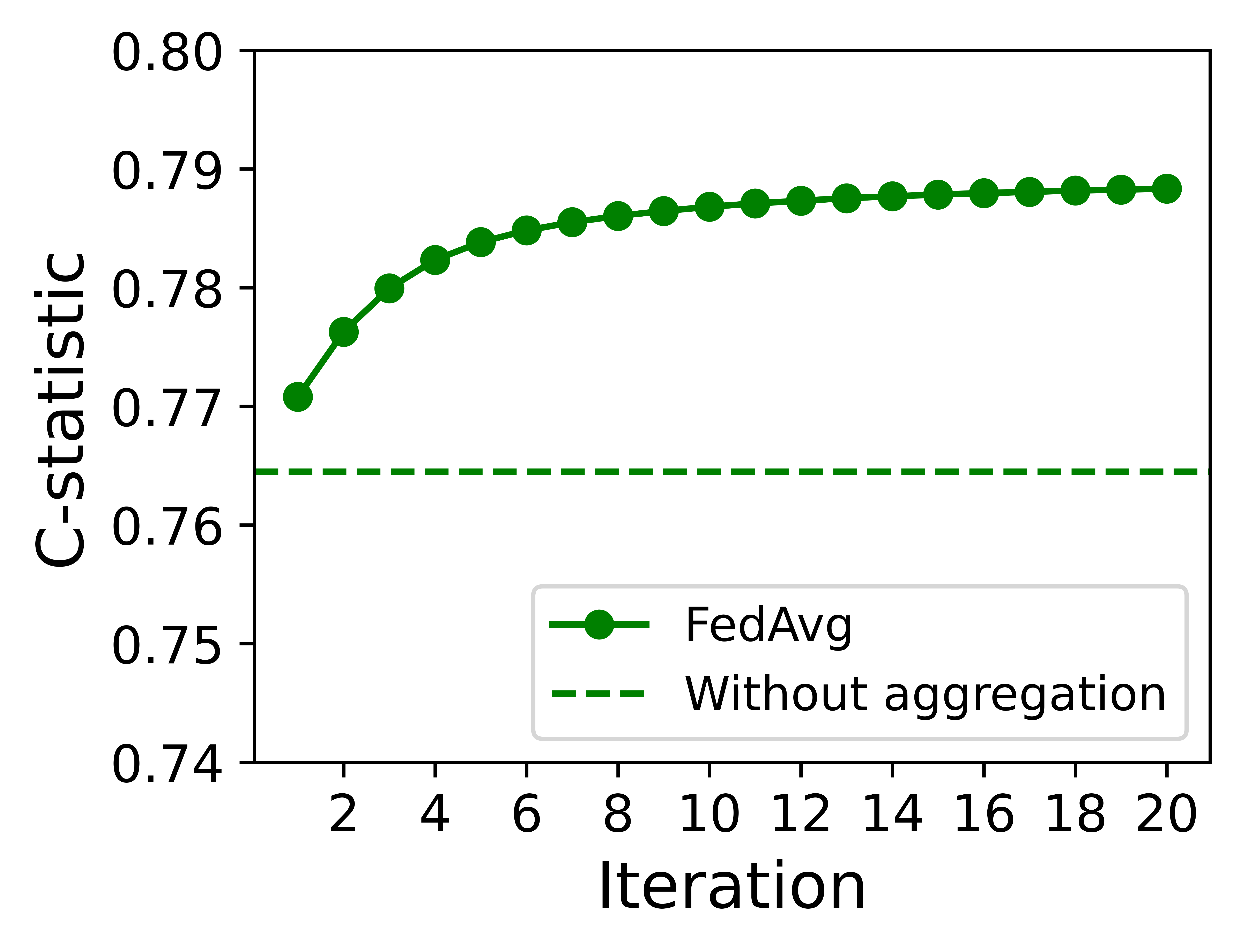}}
 \qquad
 \subfloat[\label{fig:fl_local_results_lifelines}]{%
 \includegraphics[width=0.44\linewidth]{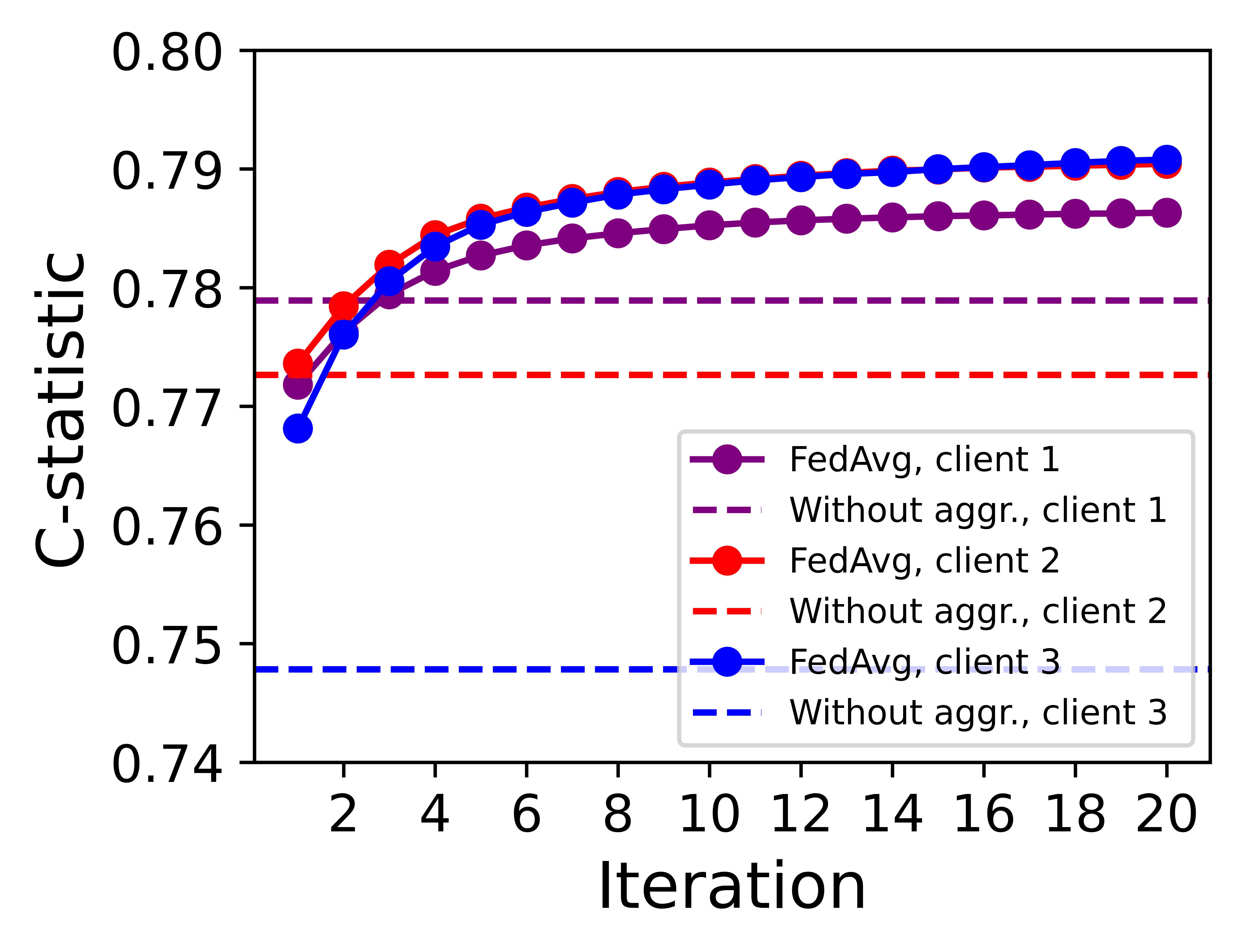}} 
 \caption{C-statistic curve presenting the discriminative ability of the proposed method, \emph{FedAvg}, compared to \emph{Without aggregation} on the Lifelines dataset (a) the global model evaluation (b) the local model evaluation.}
 \label{fig:fl_results_lifelines} 
\end{figure}

To highlight the improvement by FL, we conduct a comparative analysis comparing the discriminative ability of FedAvg with the evaluation of the local models without aggregation. Specifically, \emph{Without aggregation} indicates the evaluation of the performance of each local model on its test set after its first local training (i.e., the evaluation between line \ref{line:local_update} and \ref{line:return} of \Cref{alg:fedavg}  for the first iteration); in this compared method, the model locally trained once and its weight never be sent to the server. Namely, in this experimental setup, the nodes are completely isolated; the model weight, as well as the data, never goes out or comes in. Although the dashed line shown in \Cref{fig:fl_results_whas} and \Cref{fig:fl_results_lifelines} spans the horizontal axis, we trained and tested the model only once and did not repeat model training on the same data for update iterations of FedAvg. Since the training loss has already been converged in each node, the presented results indicate the best performance we could get on the test data in the isolated node. \par
In our experiments, a stop criterion of the proposed method is the fixed number of maximum update iterations $I$=20, which is enough to show performance improvement. In applications after this PoC, the stop criterion can be no improvement (convergence) in global model performance instead of the predetermined number of iterations. For all the experimental results, we reported the mean performance with ninety-five percent confidence intervals ($95\%$CI) through the corrected resampled t-test \cite{nadeau1999inference} based on 10 independent runs with a random split. \par

\begin{table}[ht!]
\footnotesize
\begin{center}
\caption{Results of the validation on the 148,230 subjects of the Lifelines dataset, in terms of C-statistic with 95$\%$ CIs through a corrected resample t-test based on 10 independent runs.}\label{tab:lifelines_results}
\begin{tabular}{l>{\centering}p{4cm}>{\centering\arraybackslash}p{4cm}}
\toprule[1pt] 
\multirow{2}*{Model evaluation}& \multicolumn{2}{c}{C statistic (95\%CI)}\\
& Without aggregation & FedAvg (after 20 updates) \\
\midrule[0.5pt]
Global & 0.764 (0.757-0.772) & 0.788 (0.781-0.796)  \\
Local (on client1) & 0.779 (0.755-0.802) & 0.786 (0.765-0.808)   \\
Local (on client2) & 0.773 (0.751-0.795) & 0.790 (0.771-0.810)   \\
Local (on client3) & 0.748 (0.717-0.778) & 0.791 (0.761-0.820)   \\
\arrayrulecolor{black}\bottomrule[1pt]
\noalign{\vskip 0.3mm} 
\end{tabular}
\end{center}
\end{table}

\Cref{tab:whas_results} presents the evaluation results on the WHAS dataset and \Cref{fig:fl_results_whas} illustrates them. As specified in \Cref{sssec:datasets}, the number of test samples is 164 for client 1, 98 for client 2, and 65 for client 3; the global model evaluation is based on the prediction results of 327 subjects. Without aggregation, Its mean result measure is 0.755. Federated learning gradually improves the discriminative ability (C-statistic improved to 0.775) for the fixed update iterations, as shown in \Cref{fig:fl_results_whas} (a). The results of the local model performance evaluation also show an improvement across aggregations, as depicted in \Cref{fig:fl_results_whas} (b). \par
The validation results on the Lifelines data are summarized in \Cref{tab:lifelines_results} with their graphical representation in \Cref{fig:fl_results_lifelines}. In our experimental setup, the data are partitioned as specified in \Cref{sssec:datasets} and dispatched to each client. We tested the local models on 14,823 subjects in client 1, 8,894 in client 2, and 5,930 in client 3. Overall, 29,647 subjects were used to evaluate the global model evaluation. Our federated learning improves the discriminative ability (C-statistic improved from 0.764 to 0.788); the mean performance of the global model is 0.764 before aggregation; it increases monotonically with each iteration of the aggregation and reaches 0.788, as illustrated in \Cref{fig:fl_results_lifelines} (a). The gradual improvement is also valid in the results of the local model performance evaluation. 
In particular, there is a large improvement in the discriminative ability of the model in client 3, and the mean C-statistic of the other clients also increases monotonically. These results verify that the proposed method could improve the discrimination performance of all the local prediction models by aggregating them. \par

\section{Discussion}\label{sec:idiscussion}




This paper introduces MyDigiTwin, a novel framework designed to address key challenges in \textit{health-digital-twin}-based cardiovascular risk prediction and proactive health management, including the need for patient engagement at scale, data fragmentation, and the privacy and security constraints associated with these sensitive health data. The following discusses these challenges, as presented in the existing literature on \textit{health digital twins}, and summarizes how MyDigiTwin contributes to overcoming them.

\subsection{\textit{Health digital twins}: adoption challenges and MyDigiTwin}

Previous works in cardiovascular medicine based on the concept of \textit{health digital twin} often rely on static population-level simulations, limiting their ability to provide personalized insight~\cite{coorey2022health, sel2024building}. Furthermore, these models rarely incorporate continuous patient input, resulting in limited patient engagement. MyDigiTwin addresses this by embedding predictive models directly into \textit{personal health environments}, empowering patients to explore hypothetical scenarios, such as the impact of lifestyle changes on their cardiovascular health.

Data disparity, as discussed in~\cite{venkatesh2024health}, is another significant obstacle to the widespread adoption of these \textit{health digital twins}, highlighting the critical need for robust data integration frameworks. Similarly, \cite{coorey2022health} and \cite{sel2024building} emphasize the complexity of incorporating heterogeneous data from various sources, underscoring the importance of seamless harmonization for effective cardiovascular modeling. To address this, the MyDigiTwin framework introduces a robust data harmonization process that can be reused across multiple reference datasets. By aligning this process with the ZIB-FHIR profile - the standard used by \textit{personal health environments} for data exchange with healthcare providers - the framework ensures not only consistency across datasets, but also provides clear semantics on any data extracted from them (units, coding systems, etc.). This enables the unambiguous identification of the specific inputs that these health environments must provide when using a model derived from these datasets, enhancing transparency and reliability in data-driven analyzes.

Equally important are privacy and security concerns, which have been consistently cited as the main barriers to the adoption of \textit{health digital twins} in healthcare~\cite{venkatesh2024health, coorey2022health}. Traditional methods often involve the sharing of raw data between institutions, which risks patient confidentiality and regulatory compliance. To address these concerns, MyDigiTwin establishes a research environment based on a federated infrastructure, allowing predictive models to be trained without sharing sensitive patient data. Instead, the data (after being harmonized) remain securely within its original location and only aggregated model weights are shared. This approach ensures compliance with privacy regulations, such as GDPR, while also facilitating collaboration among multiple institutions to develop robust and accurate models. This aligns with the recommendations of \cite{venkatesh2024health}, advocating for privacy-preserving techniques, and marks a critical advance in utilizing big data for personalized care without compromising patient privacy.

\subsection{Insights from the Proof of Concept}

The proof of concept described in Section~\ref{sec:poc} demonstrated that the MyDigiTwin framework can support the development of CVD prediction models using consistent and semantically rich variables from multiple reference datasets through the harmonization pipeline detailed in \Cref{sec:harmonization}. This prepares for the implementation of the actual prediction models once available, as these are currently being developed in the context of the MyDigiTwin project.


\par The process followed in the creation of the model described in Section~\ref{sec:poc} not only illustrates the end-to-end integration, from raw data to a model based on previously identified cardiovascular disease (CVD) predictors, but also highlights how the incorporation of additional harmonized data sets can enhance model performance. This approach ensures that predictive models are built on robust and standardized data, facilitating accuracy. \par 
The experimental results reported in \Cref{ssec:validation} support the validity of the proposed system architecture of the MyDigiTwin framework (\Cref{fig:fhir-for-training-prediction-consistency}). While the data in the nodes have not been globally visible and the prediction model has been trained locally in each client, our federated learning (i.e., the aggregation of model weights) did not decrease the discriminative ability of the model's predictions at any node. Also, we could observe the convergence of the global model performance after monotonic improvement across aggregation iterations. Hence, our validation results on the Lifelines dataset show that it is feasible to use large-scale federated data for predictive modeling in our MyDigiTwin framework. \par

\subsection{Limitations and challenges}

As discussed in Section~\ref{sec:poc}, due to the ongoing process of acquiring new reference datasets, a simulated federated setting was created by randomly splitting the harmonized Lifelines dataset across multiple nodes. However, in a real-world scenario that involves data from diverse sources, challenges extend beyond data format and semantic heterogeneity. These settings also introduce the complexities of statistical heterogeneity, as noted by~\cite{ye2023heterogeneous}.

Additionally, preprocessing tasks such as data imputation and normalization, which are relatively straightforward when working with centralized data, become significantly more complex in federated learning due to the distributed nature of the data. In a federated environment, data remains at its original location, which complicates the determination of optimal imputation and normalization strategies, as the researcher lacks global visibility of the entire dataset. Consequently, these preprocessing tasks become privacy-preserving data processing problems in their own right, in addition to the primary research task at hand~\cite{du2022rethinking, oishi2023federated}.

Another significant challenge lies in the envisioned end-user environment, where patients will access predictive and simulation services via the \textit{personal health environment}. The MyDigiTwin module, which connects pre-trained prediction models with the services of \textit{personal health environments}, is expected to be developed in collaboration with the MedMij Foundation. While the adoption of FHIR standards is expected to facilitate technical integration, barriers related to technology access and digital literacy, particularly among underserved populations, could hinder the widespread use of \textit{personal health environments} in general and platforms like MyDigiTwin in particular. Ensuring accessibility and usability for diverse patient groups is crucial for the framework’s success and remains an area for ongoing improvement.

\section{Conclusions and future work}\label{sec:conclusions}

MyDigiTwin is a novel framework aimed at advancing primary prevention in cardiovascular healthcare. By integrating personal risk assessment features of \textit{health digital twins} with \textit{personal health environments}, it empowers patients while ensuring the privacy of health records, as no data is exchanged outside this \textit{health environment} infrastructure. Population-based insights are leveraged to enable patients to explore personalized health scenarios, with predictive models trained on harmonized datasets within a federated research environment. This environment, supported by a data harmonization framework, ensures semantic consistency across diverse datasets, enabling more accurate and representative predictions.

The proof-of-concept demonstrated the feasibility of harmonizing diverse big-data reference datasets and their integration with health records for training predictive models. These models enable personalized scenario exploration, motivating patients toward healthier lifestyle changes and supporting primary prevention.

Future work will focus on implementing this framework in real-world settings. This approach has the potential to shift from reactive care, where patients are treated after symptoms appear toward proactive strategies, enabling early detection and prevention of cardiovascular disease. Additionally, the proposed architecture could contribute to uncovering novel CVD predictors, enhancing predictive models to improve patient outcomes, and incorporating new data types such as computed tomography scan images, to broaden its applicability.

An equally important area of future research should focus on fostering trust and understanding between clinicians and patients. Enhancing transparency in AI decision-making and developing intuitive tools for cardiovascular disease management will allow clinicians to effectively interpret model outputs and communicate these findings to patients. This collaborative approach will enable patients to better understand the reasoning behind proposed treatments, strengthening confidence and promoting shared decision-making in the application of digital twin models.

\section*{Ethics statement}

The manuscript is based on data from the Lifelines Cohort Study and the Worchester Heart Attack Study. The original Lifelines protocol was approved by the UMCG Medical ethical committee under reference number 2007/152 and all participants provided informed consent. This study was approved by the Lifelines research office under reference number OV22-0581. The Worchester Heart Attack Study is publicly available at the website\footnote{\url{https://web.archive.org/web/20170114043458/http://www.umass.edu/statdata/statdata/data/}} managed by University of Massachusetts, which provides a collection of datasets for educational purposes.  All data were processed in compliance with the General Data Protection Regulation. 

\section*{Declaration of generative AI and AI-assisted technologies in the writing process}

During the preparation of this work, the author(s) used AI-assisted technologies only to improve the readability and language of the manuscript. Any text generated by these tools was thoroughly reviewed and edited by the authors.

\section*{Acknowledgments}

This publication is part of the project MyDigiTwin with project number 628.011.213 of the research programme ``COMMIT2DATA – Big Data \& Health'' which is partly financed by the Dutch Research Council (NWO). Furthermore, this work used the Dutch national e-infrastructure with the support of the SURF Cooperative using grant no. EINF-7675.
We thank the MyDigiTwin consortium members for the useful and inspiring discussions.

\bibliographystyle{cas-model2-names}

\bibliography{cas-refs,tools,related_works}


\end{document}